\documentclass[letterpaper, 10 pt, conference]{ieeeconf}  

\IEEEoverridecommandlockouts                              

\title{\LARGE \bf
\vspace{-7.5mm}
Towards a Causal Probabilistic Framework for Prediction, Action-Selection \& Explanations for Robot Block-Stacking Tasks
}

\author{Ricardo Cannizzaro$^{*}$, Jonathan Routley$^{*}$, and Lars Kunze
\thanks{All authors are with the Oxford Robotics Institute,
Department of Engineering Science, University of Oxford, UK.}%
\thanks{This work is supported by Defence Science \& Technology Australia and the EPSRC RAILS project (grant reference: EP/W011344/1).}%
\thanks{* Cannizzaro and Routley are identified as joint lead authors of this work.}
}

\usepackage{mathtools}

\usepackage{csquotes}
\usepackage{xcolor}

\usepackage{tikz}
\usetikzlibrary{shapes,decorations,arrows,calc,arrows.meta,fit,positioning}
\tikzset{
    -Latex,auto,node distance =1 cm and 1 cm,semithick,
    state/.style ={ellipse, draw, minimum width = 0.8 cm, minimum height=0.8cm},
    point/.style = {circle, draw, inner sep=0.04cm,fill,node contents={}},
    bidirected/.style={Latex-Latex,dashed},
    el/.style = {inner sep=2pt, align   =left, sloped}
}
\tikzstyle{arrow} = [thick,->,>=stealth]

\begin{document}

\maketitle
\thispagestyle{empty}
\pagestyle{empty}


\begin{abstract}
Uncertainties in the real world mean that is impossible for system designers to anticipate and explicitly design for all scenarios that a robot might encounter. Thus, robots designed like this are fragile and fail outside of highly-controlled environments. 
Causal models provide a principled framework to encode formal knowledge of the causal relationships that govern the robot's interaction with its environment, in addition to probabilistic representations of noise and uncertainty typically encountered by real-world robots.
Combined with causal inference, these models permit an autonomous agent to understand, reason about, and explain its environment. 
In this work, we focus on the problem of a robot block-stacking task due to the fundamental perception and manipulation capabilities it demonstrates, required by many applications including warehouse logistics and domestic human support robotics.
We propose a novel causal probabilistic framework to embed a physics simulation capability into a structural causal model to permit robots to perceive and assess the current state of a block-stacking task, reason about the next-best action from placement candidates, and generate post-hoc counterfactual explanations.
We provide exemplar next-best action selection results and outline planned experimentation in simulated and real-world robot block-stacking tasks.
\end{abstract}


\section{INTRODUCTION}
Previous work has found that human judgements about the stability of 3D block towers correlate closely to a predictive model that uses physics simulations to forward-simulate future outcomes given noisy beliefs about the current world state \cite{hamrick2011internal}. 
In this work we aim to provide robots the ability to perform this kind of \enquote*{mental} simulation, to thus endow them with an intuitive understanding of the physics that govern their operating environment,
a capability identified as a crucial component to build systems that are able to think and reason robustly in a similar manner to humans \cite{Lake2017BuildingPeople}.

However, this understanding of physics of an autonomous agent is predicated on the possession of a model that encodes knowledge of the causal relationships that exist between entities in the world, the robot and its task, as well as the uncertainties in the robot and environment.
Crucially, the use of a causal model permits an autonomous agent to utilise association, interventional, and counterfactual inference, which are foundational components of agents to think, plan, and explain.
We propose a novel probabilistic causal framework and to demonstrate that combining 3D physics simulations with causal modelling and inference, allows a robot to make accurate predictions, decisions, and explanations of observed outcomes (e.g., successes and failures) in a manner that is robust to various sources of noise and uncertainty.

The three main contributions of this work are: 1) the description of a probabilistic structural causal model (SCM) of a robot block-stacking task; 2) a proposal of a causal modelling and inference framework for reasoning about the block-stacking task, including stability prediction, next-best block placement action-selection, and generation of counterfactual explanations to identify the cause of observed task outcomes in a human-understandable manner; and, 3) demonstrate with initial results the effectiveness of the proposed framework and planned experimentation on a simulated and real-world robot.


\begin{figure}[t]
    \centering
    \includegraphics[width=0.95\linewidth]{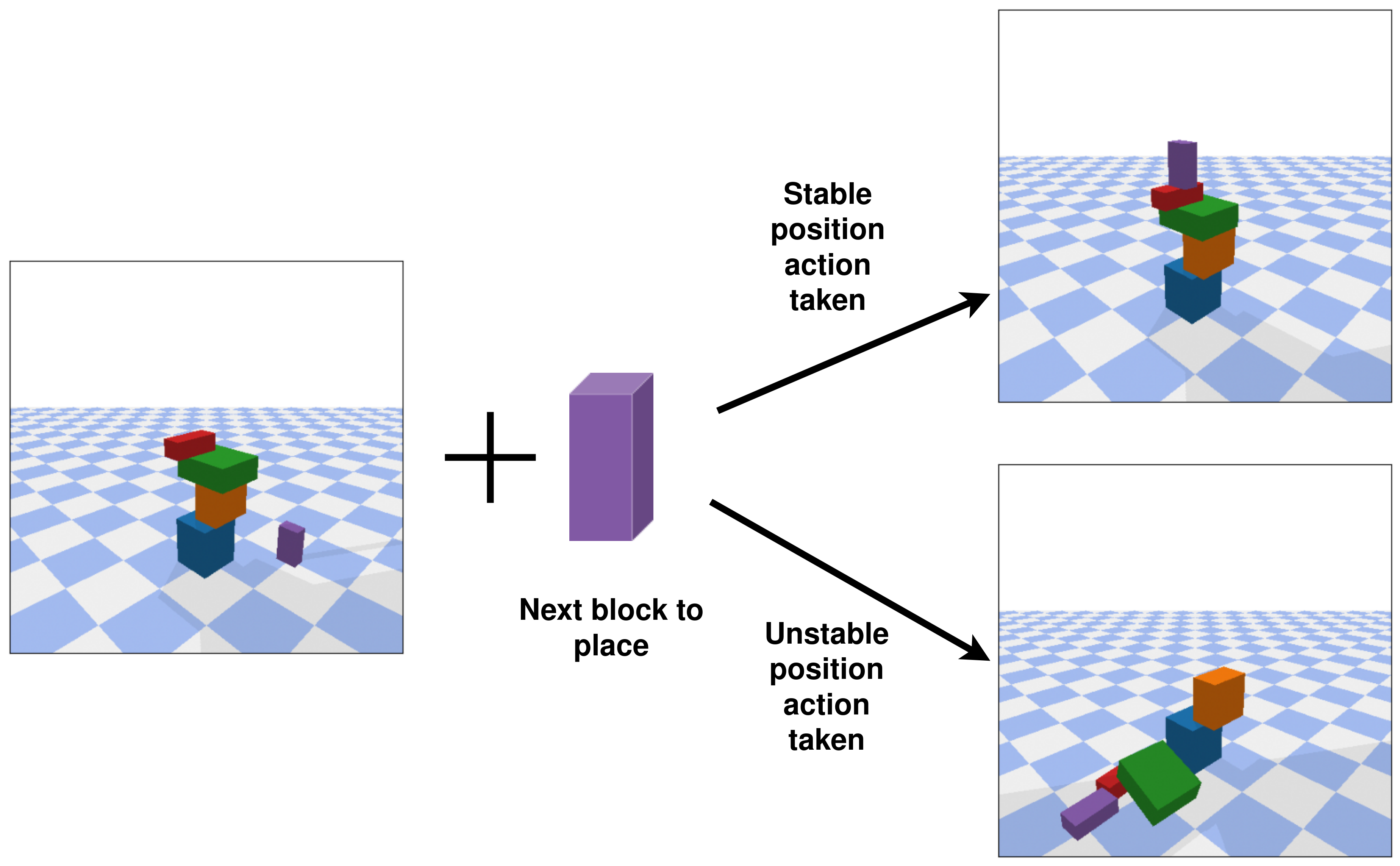}
    \caption{Possible action-selection outcomes for an initially stable tower. The initial state $S_0$ and the next block to be added are shown, along with the final state $S_1$ after adding the block to a stable and unstable position.}
    \label{fig:problem-decision}
\end{figure}
\section{PROBLEM DESCRIPTION} \label{sec:problem_definition}
%

            
%
\subsection{Robot Block-Stacking Task}
In the sequential block-stacking task considered here (Fig. \ref{fig:problem-decision}), the robot's task is to incrementally build a tower from an initial configuration of one or more blocks by stacking a defined sequence of blocks, each with specified mass, dimension, and colour properties, such that after each individual block placement the tower remains standing without any external support. 
The task is considered a failure if the tower falls at any point, otherwise a success if the tower remains standing after the final block placement.
A key challenge is the presence of noise and errors typically experienced by robot sensors, estimation, and control; these uncertainties must be considered for robots to complete the task reliably.

To avoid the computational challenges of sequential decision-making problems, we consider only single-column towers and constrain the task to a sequence of independent next-best action-selection problems \textemdash\ i.e., we plan only one move ahead. 
We formulate each next-best block placement problem as finding the optimal action $\hat{a}$ from the  action set $A$, parameterised by block placement position, that gives the maximum expected probability of the resultant tower being stable: $\hat{a} = \underset{a \in A}{\mathrm{argmax}} \{ P(Stable=True \ | \ Action=a ) \}$.
\subsection{Post-Hoc Counterfactual Explanations}
We also address the problem of creating a capability for the robot system to create intuitive explanations for sensor measurements, action-selections, and observed task outcomes, following a block-stacking task attempt.
\section{RELATED WORK}
Robot manipulation is a long-standing field of research. However, only recently has causality been leveraged for robot manipulation tasks. In recent work by Diehl \& Ramirez-Amaro \cite{diehl2022causalbased}, authors model the block-stacking task as a causal Bayesian network (CBN). They learn task success probabilities from physics-based simulation data and demonstrate action selection using the learned model. However, since the model is fixed after training, any change to block attributes would require the model to be re-trained, limiting its generalisability. Additionally, the robot task uncertainties are not explicitly included as random variables, limiting the interpretability of the model.
Further, they use the CBN in a breadth-first-search algorithm to generate contrastive explanations; however, these may diverge from counterfactual explanations when the outcome is sufficiently impacted by random variables. 
\section{PROPOSED FRAMEWORK}
\subsection{SCM Model Formulation}
We propose an SCM formulation of the robot block-stacking task, with an underlying causal directed acyclic graph (DAG) that: 1) explicitly models uncertainty in perception and actuation to permit reasoning over different sources of uncertainty; and, 2) uses a physics-based simulator online at model execution-time to allow the model to generalise and scale well. Specifically, we adopt a SCM model representation to permit counterfactual explanations.

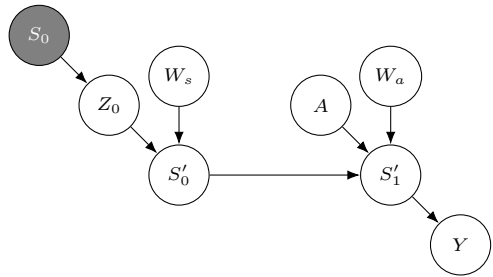
\begin{figure}[t]
    \scriptsize
    \centering
    \begin{tikzpicture}[node distance=0.5cm]
        \node[state] (s-prime-zero) at (0,0) {$S'_0$};
        \node[state] (s-noise) [above = of s-prime-zero, ] {$W_s$};
        \node[state] (z-zero) [above left = of s-prime-zero, ] {$Z_0$};
        \node[state] (s-zero) [above left = of z-zero, fill=gray, text=white] {$S_0$};
        \node[state] (s-prime-one) [right = of s-prime-zero, xshift=1.5cm] {$S'_1$};
        \node[state] (a-noise) [above = of s-prime-one] {$W_a$};
        \node[state] (action) [above left = of s-prime-one] {$A$};
        \node[state] (outcome) [below right= of s-prime-one] {$Y$};

        \path (s-zero) edge (z-zero);
        \path (z-zero) edge (s-prime-zero);
        \path (s-noise) edge (s-prime-zero);
        \path (s-prime-zero) edge (s-prime-one);
        \path (a-noise) edge (s-prime-one);
        \path (action) edge (s-prime-one);
        \path (s-prime-one) edge (outcome);
    \end{tikzpicture}
    \caption{The proposed model of the block-stacking task, shown as a causal DAG. White nodes represent observable variables, grey unobservable.}
    \label{fig:model-causal-dag}
\end{figure}

The data generation process for the system is modelled by the DAG shown in Fig. \ref{fig:model-causal-dag}. 
$S_0$ represents the hidden true state of the tower.
The robot draws an observation $Z_0$ from $S_0$, containing additive sensor noise parameterised by $W_s$, which is used to form the robot's belief about the tower state $S'_0$. 
The resultant state of the tower is sampled from the transition function $T(s'_0,a,w_a,s'_1 )$, specifying the probability distribution over arriving in successor state $s'_1$ after taking action $a$ from current state $s'_0$, with actuation noise $w_a$ sampled from the action distribution $W_a$. This transition function sampling is performed online using a 3D physics simulator.
%
\subsection{Stability Prediction \& Action Selection}
\begin{figure*}[t]
      \centering
      \includegraphics[width=0.75\linewidth]{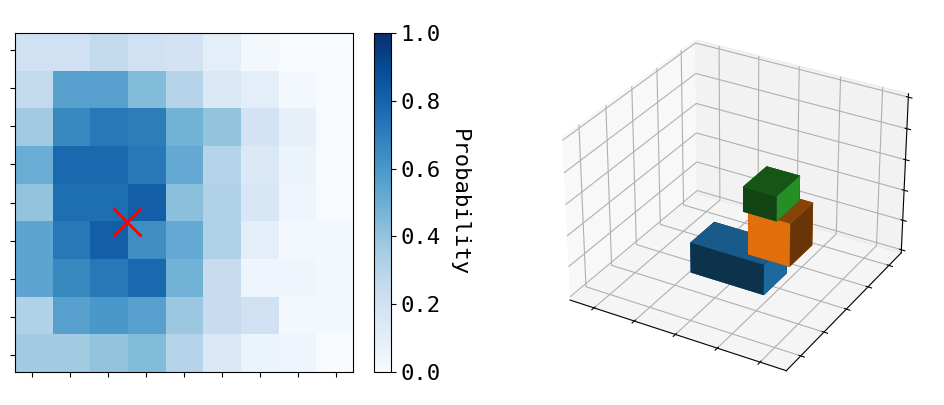}
      \caption{An illustration of the next-best action-selection process. LHS: A heatmap of inferred tower stability probabilities over candidate positions, with the selected position shown as a red cross. RHS: Initial tower state.}
      \label{fig:stability-heatmap}
      \vspace{5mm}
\end{figure*}
Given our proposed causal model formulation, we can predict the probability of a given tower state being stable following a candidate block placement action $a$ by inferring the posterior distribution: $P\big(IsStable(s_1')|s'_0,do(A=a)\big)$. Here, the agent's choice of action is represented by an intervention on the model, using the $do(\cdot)$ operator. For stability prediction of an initial tower state without the robot placing another block, we simply formulate the inference query with the robot agent taking a NULL action (i.e., $do(A=NULL)$) \textemdash\ this formulation allows us to re-use the same model for stability prediction and action selection. To solve the next-best action-selection problem, we perform a uniform sampling of candidate block placement positions over the top surface area of the current top block, find the subset of positions with expected stability over a given threshold, and select the geometric mean of these subset positions as the choice of block placement. Figure \ref{fig:stability-heatmap} illustrates this process on an example initial tower state.
\subsection{Post-Hoc Counterfactual Explanations}
Counterfactual inference permits an agent to simulate what would have happened, had certain things in past been different to the observed reality. 
Our proposed approach to generating post-hoc counterfactual explanations of an observed block-stacking task execution will use twin-world counterfactual algorithm based methods to search for the random variable with the highest probability of being a necessary and sufficient cause for the observed outcome. 
Explanation variables will include the initial state, sensor and actuator noise variables, and the agent's selected action.
%
%
\section{INITIAL AND PLANNED EXPERIMENTATION}
\begin{figure*}[t]
      \centering
      \includegraphics[width=0.95\textwidth]{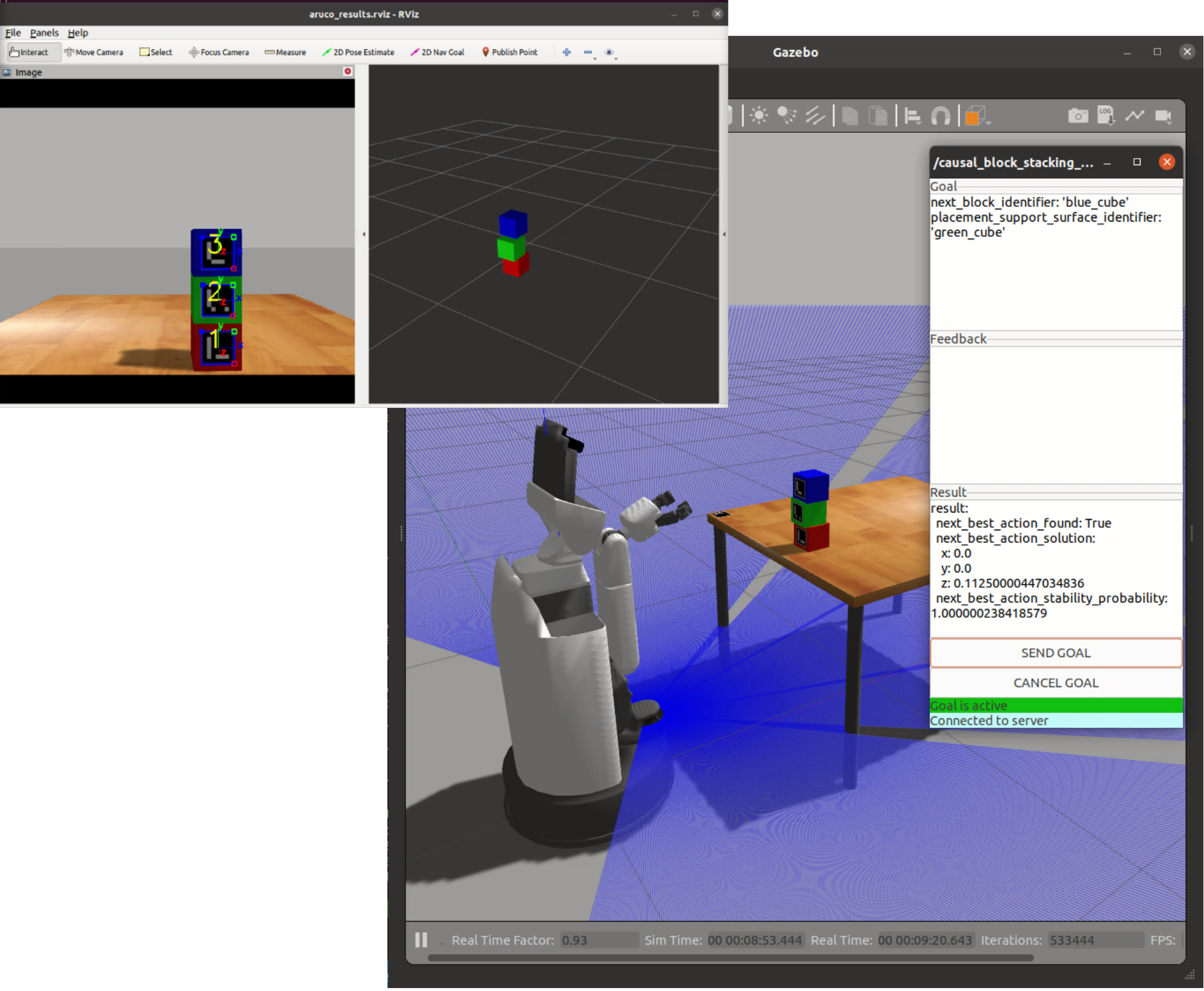}
      \caption{An illustration of the ROS-based next-best action-selection functionality executed by the HSR robot model in the Gazebo physics-based simulation environment. Top left-hand corner: RVIZ visualisation showing the image from the Gazebo-simulated camera with Aruco marker detection annotations (left) and subsequent 3D block tower state reconstruction (right). Right-hand side: Gazebo simulation world including the HSR model, table, and three coloured blocks stacked in their final configuration following a manipulation action to add the blue block to the tower. A simple ROS action client is also shown, demonstrating an exemplar call to the next-best action-selection ROS Action server and returned inference and decision-making result.}
      \label{fig:gazebo-sim}
\end{figure*}

We plan to evaluate the performance of our proposed framework, integrated with a Toyota Human Support Robot (HSR) in simulation-based and real-world experimentation. 
The framework will be interfaced with via ROS and integrated with the existing HSR perception and manipulation sub-systems.
ROS-based tower stability prediction and next-best action selection using perception inputs are already integrated and operating in simulation at the time of writing. Figure \ref{fig:gazebo-sim} illustrates the developed ROS-based next-best action-selection functionality executed by the HSR robot model in the Gazebo physics-based simulation environment.

We will consider metrics including the prediction accuracy compared to real-world action success rates, optimality of action-selection compared to other baseline policies and planners, and a qualitative evaluation of counterfactual observations as compared to human causal explanations of observed task outcomes. We seek to demonstrate the robustness of predictions due to accounting for system uncertainty, optimality of action-selection, and the interpretability and extent of agreement with human causal judgements of the generated post-hoc counterfactual explanations.

\bibliographystyle{ieeetr}
\bibliography{references.bib}

\end{document}